\documentclass[doublecol]{epl2}
\usepackage{graphicx}
\usepackage{amsmath,amssymb}

\usepackage[usenames]{xcolor}
\usepackage[normalem]{ulem}
\usepackage{url}

\let\oigr\includegraphics
\def\includegraphics[#1]#2{\IfFileExists{#2.eps}{\oigr[#1]{#2}}{\oigr[#1]{figures/#2}}}

\title{Sequencing Chess}
\author{
  Arshia Atashpendar\inst{1}
  \and
  Tanja Schilling\inst{1}
  \and
  Thomas Voigtmann\inst{2}
}
\shortauthor{A.~Atashpendar, T.~Schilling, and Th.~Voigtmann}
\institute{
\inst{1}
  Theory of Soft Condensed Matter, Universit\'e du Luxembourg, L-1511 Luxembourg, Luxembourg\\
\inst{2}
  Institut f\"ur Materialphysik im Weltraum, Deutsches Zentrum f\"ur Luft- und Raumfahrt (DLR), 51170 K\"oln, Germany\\
  Department of Physics, Heinrich-Heine-Universit\"at D\"usseldorf, 40225 D\"usseldorf, Germany
}
\begin{document}

\abstract{
We analyze the structure of the state space of chess by means of transition path sampling Monte Carlo simulation.
Based on the typical number of moves required to transpose a given
configuration of chess pieces into another, we conclude that the state space consists
of several pockets between which transitions are rare.
Skilled players explore an even smaller subset of positions that populate
some of these pockets only very sparsely.
These results suggest that the usual measures to
estimate both, the size of the state space and the size of the tree of legal
moves, are not unique indicators of the complexity of the game, but that
topological considerations are equally important.
}



\maketitle

Chess is a two-player board game with a small set of rules
according to which pieces can be moved. It belongs to the class of games
with perfect information that have not been solved yet, due to the sheer
size of its state space. 
The computerized analysis of chess started with a seminal paper by
Claude Shannon in 1950 \cite{Shannon1950}, and since about the year 2000
computer programs can regularly beat top-level human players
\cite{Campbell2002}. They do so by
employing well-tailored heuristic evaluation functions for the game's states,
which allow one to short-cut the exploration of the vast game tree of possible
moves.
In this context, chess is often compared to Go, where
computers only very recently started to match the performance of
human champions \cite{Silver2016}.
The difference is usually attributed to the different sizes of the games'
state spaces: the game-tree complexity of Go exceeds that of chess by
some 200 orders of magnitude.

However, while size is an important factor in determining the complexity of
a game, the topology of the state space may be equally important.
Intuitively, the different kinds of moves performed by different
chess pieces impose a highly nontrivial
(and directed) topology. It is not at all straightforward to establish
whether a given point in the state space is reachable from another one
by a sequence of legal moves.

We thus face an interesting sampling problem:
given two chess configurations, can one establish whether they are
connected, i.e., whether there exists a sequence of legal moves that
transforms the first configuration into the second? Furthermore, what
is the typical distance (in plies, or half moves) between such configurations?
Clearly, direct enumeration or standard Monte Carlo sampling are out
of reach: after each ply, the game tree is estimated to branch into 30 to 35
subtrees \cite{Shannon1950}.

Here we demonstrate that it is possible to analyze the topological structure
of the state space of chess by stochastic-process rare-event sampling
(SPRES) \cite{Berryman2010}. SPRES is a transition-path Monte Carlo sampling
scheme that works in full non-equilibrium conditions, where the dynamics
is neither stationary nor reversible.
\footnote{Our analysis of chess also serves to demonstrate the versatility
and power of SPRES as a technique that applies to abstract non-physical
dynamics.}
Combining SPRES with an optimized chess-move generator~\cite{Arshia}, we estimate the distribution of
path lengths between both randomly generated configurations and those
encountered in games played by humans. Analyzing these distributions in
terms of random-graph theory, we conjecture that the state space of
chess consists of multiple distinct pockets, interconnected by relatively
few paths. These pockets are only very sparsely
populated by the states that are relevant for skilled play.

Previous statistical-physics analyses of chess have focused mostly
on the distribution of moves in human gameplay,
or on games played by computer
chess engines. For example, the
popularity of opening sequences follows a power-law distribution according
to Zipf's law
\cite{Blasius2009} (in this context, Go is rather similar \cite{Xu2015}),
highly biased by the skill of the players involved
\cite{Maslov2009,Schaigorodsky2014}.
Optimal play (in the sense that moves are evaluated favorably by modern computer chess engines) has also been analyzed in the language of free-energy
landscapes \cite{Krivov2011}.
Our approach is entirely different: we consider the set of all legal
moves, irrespective of their engine evaluation, in order to establish
the connectivity of the state space of chess. Within this space, we then also study 
the relative size and structure of the subset of positions encountered in 
games played by chess masters.

The \emph{state} of a chess game at any point in time is entirely described by
the board configuration (the positions of all chess pieces), a small
set of additional variables that track the possibility of special moves (castling or
en-passant capture) and the information regarding which player's turn it is.
The set of \emph{possible} states is given by all states that involve
up to 16 chess pieces per color (there may be fewer due to captures, and the
number of pieces and pawns may change due to pawn promotions).
Only a subset of all possible states is \emph{legal}, as for example,
the two kings cannot be in check at the same time.
Of interest in the following are states that are legal and also
\emph{accessible} from the given initial
configuration.
As an example of an inaccessible but legal state, consider the case where the position of a bishop differs from its initial position, while the positions of the pawns do not. This state is inaccessible, because pawns are initially placed in front of the other pieces of their colour, their moves are always irreversible and the other pieces (apart from the knights) cannot jump over the pawns. Thus, although the state is legal, it cannot be reached by legal moves.

To sample the structure of the state space, we generate sequences of accessible states by randomly drawing moves evenly from all legal moves (Monte Carlo, MC).
Most of these states entail dramatic disadvantages for at least one side.
Therefore, the set of states encountered in optimal-strategy play is
vastly smaller than the set we sample. As a proxy to these unknown optimal states, we
use \emph{database} (DB) states extracted from a database of about two million
human-played games \cite{twic}.
In both cases (MC and DB), we then pick pairs of states randomly and establish their connectivity with respect to the game tree by all legal (MC) moves, i.e., irrespective of 
whether the connecting pathway contains unfavorable positions in terms of gameplay.

In the vicinity of the starting configuration, many randomly drawn pairs of positions are
necessarily disconnected, since pawns only move forward and many of the
pieces still have to gain freedom to move. At the other end of the game, mating
positions act as absorbing states. And in addition, the MC dynamics
has a set of absorbing states where only the kings are left on the board.

In order to sample states that reflect the intrinsic topology of the
state space, we thus restrict the discussion to pairs of states drawn from
a depth between 5 and 50 plies into the game. This corresponds loosely
to chess players' notion of the middle game.
Inside this window, we did not find an obvious correlation between the 
ply-depth from which a pair of states was drawn and the separation between them.

We sample the pathways between states by means of SPRES \cite{Berryman2010}.
In this method, interfaces in state space are defined by 
constant values of a scalar reaction coordinate, which quantifies the progress made from one state to the other. Then adaptive sampling of dynamic pathways is carried out such that a constant 
number of forward transitions between these interfaces is obtained. Once the sampling is completed,
observables can be averaged over the ensemble of sampled pathways. In the case of chess, we are in particular interested in the length (number of plies) of the shortest path between two configurations.

While the choice of an optimal reaction coordinate is a topic in its own right \cite{Krivov2011},
we make use of the fact that SPRES will sample paths faithfully even
for non-optimal choices \cite{Berryman2010}. 
As the reaction coordinate, we chose
a Euclidean geometric measure of distance from the target configuration.
For each piece, the geometric distance is calculated using a metric
that is adapted to the type of moves performed by that piece:
Chebyshev metric for queens, kings, and bishops, the ceil of half
the Chebyshev distance for knights, the Manhattan distance for rooks,
and the rank separation for pawns. (For details, see Ref.~\cite{Arshia}).
Pairs are discarded as disconnected if they are farther apart than 120 plies;
this approximation is adapted to the typical length of real chess games.
Trivially disconnected pairs are discarded by an initial test based on
the reaction coordinate, the pawn structure and the piece count.
For the estimation of path lengths, 4000 (3000) pairs generated from MC (DB)
that have passed this test have been sampled.


\begin{figure}
\begin{center}
\includegraphics[width=1.0\linewidth]{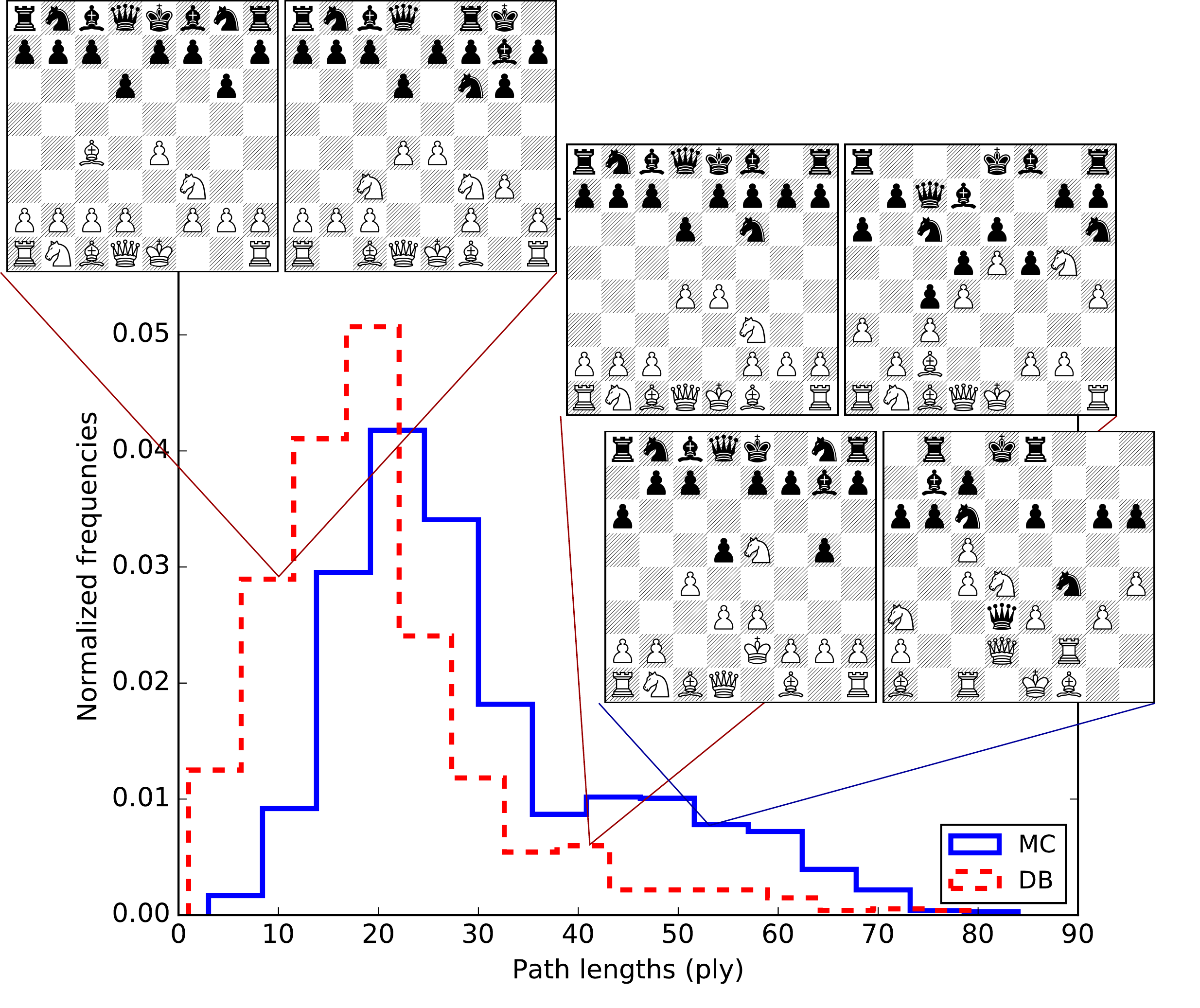}
\end{center}
\caption{\label{fig:histogram}
  Distribution of path lengths between randomly selected pairs of chess
  states as found by SPRES sampling. Pairs are drawn from a
  database of real games (DB, red dashed line), respectively
  generated via Monte Carlo dynamics (MC, blue solid line).
  In each case, sampling was restricted to starting and ending
  states between 5 and 50 plies into the game.
  Three pairs of configurations (two for DB, one for MC) are shown as
  examples connected by lines to the corresponding bins in the histograms.
  In each example, black is to move first, the left board shows the starting
   configuration, and the right board the target configuration
  \protect\cite{config1}.
}
\end{figure}

Figure~\ref{fig:histogram} shows the histogram of path lengths between
those randomly chosen pairs that are connected according to
SPRES (corresponding to $79\%$ of all randomly drawn MC pairs and $85\%$ of all
pairs drawn from the DB). 
For pairs generated via MC, the path-length distribution has two distinct
contributions, one with a peak at $\ell_1\approx20\,\text{plies}$, and a smaller one
at $\ell_2\approx45\,\text{plies}$.
The path-length distribution between pairs sampled from the database
is biased to smaller path lengths and
has only one prominent peak at a path length slightly below $\ell_1$,
$\ell_1'\approx18\,\text{plies}$.
A tail towards large distances is still seen as a remnant of the second
peak found in the MC distribution.
Note that the paths found by SPRES for the DB pairs almost certainly
pass through non-DB states (i.e.~states that are usually not found in games played by humans). A typical 
engine evaluation function (Stockfish \cite{stockfish}) displays huge fluctuations along the SPRES paths,
indicating that these paths will probably never be chosen by skilled human players.

\begin{figure}
\begin{center}
\includegraphics[width=0.95\linewidth]{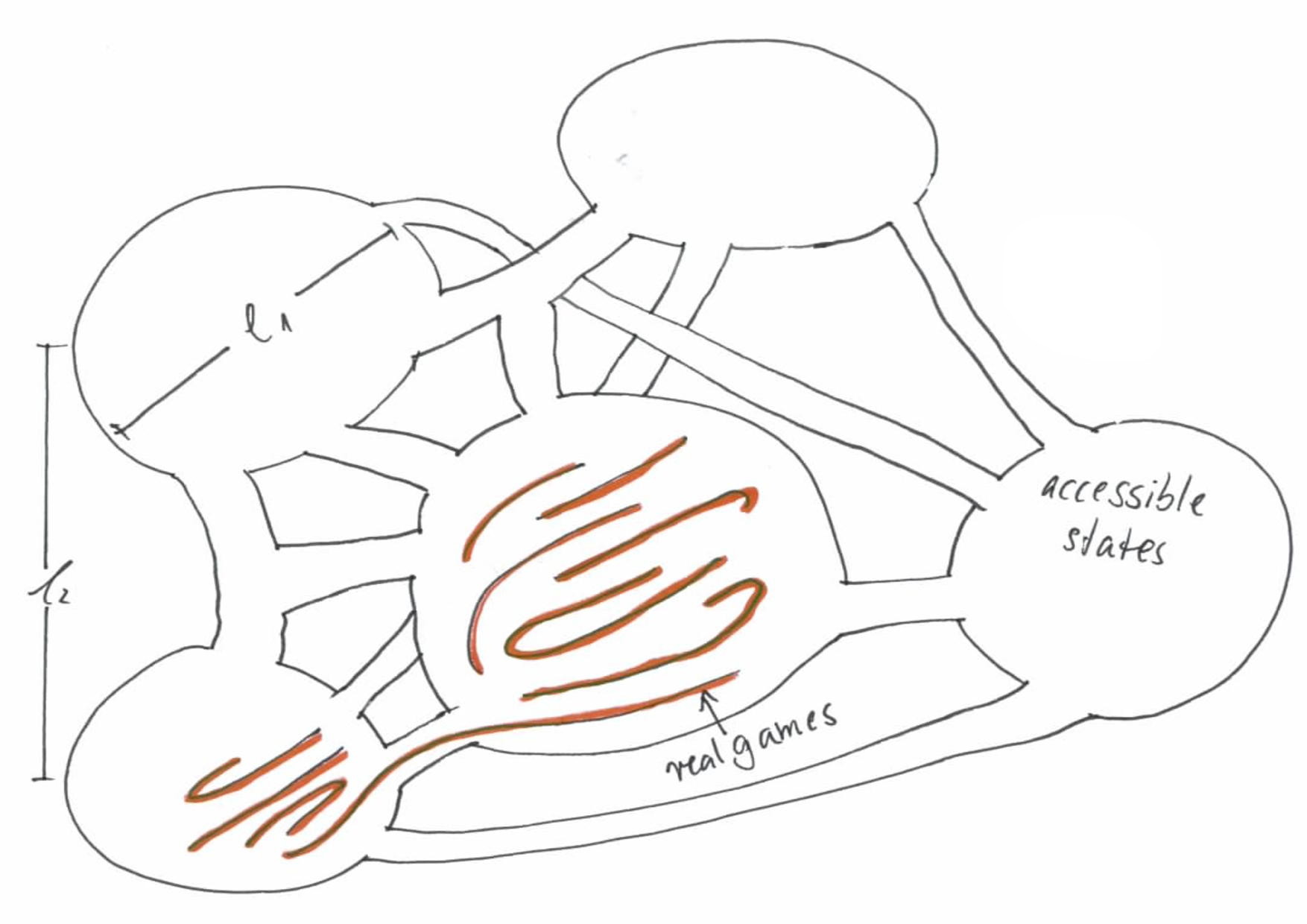}
\end{center}
\caption{\label{fig:sketch}
  Sketch of the structure of the state space of chess.
}
\end{figure}

The results shown in Fig.~\ref{fig:histogram} reveal that real chess games
take place in a subspace that is much more tightly connected than the
space of accessible states. The double-peaked histogram
suggests a ``blob'' structure (see sketch in Fig.~\ref{fig:sketch}): the space of 
accessible states consists of
pockets with average distances $\ell_1\sim20$ between nodes, and real games
are embedded in these pockets. The pockets are interconnected by long paths
of $\ell_2\sim45$, and most of them are devoid of real-game configurations.
Path lengths sampled between one MC and one DB state follow a histogram
similar to the one shown for MC pairs in the figure, confirming that the DB states
are indeed part of the state space sampled in our MC dynamics.

\begin{figure}
\begin{center}
\includegraphics[width=1.0\linewidth]{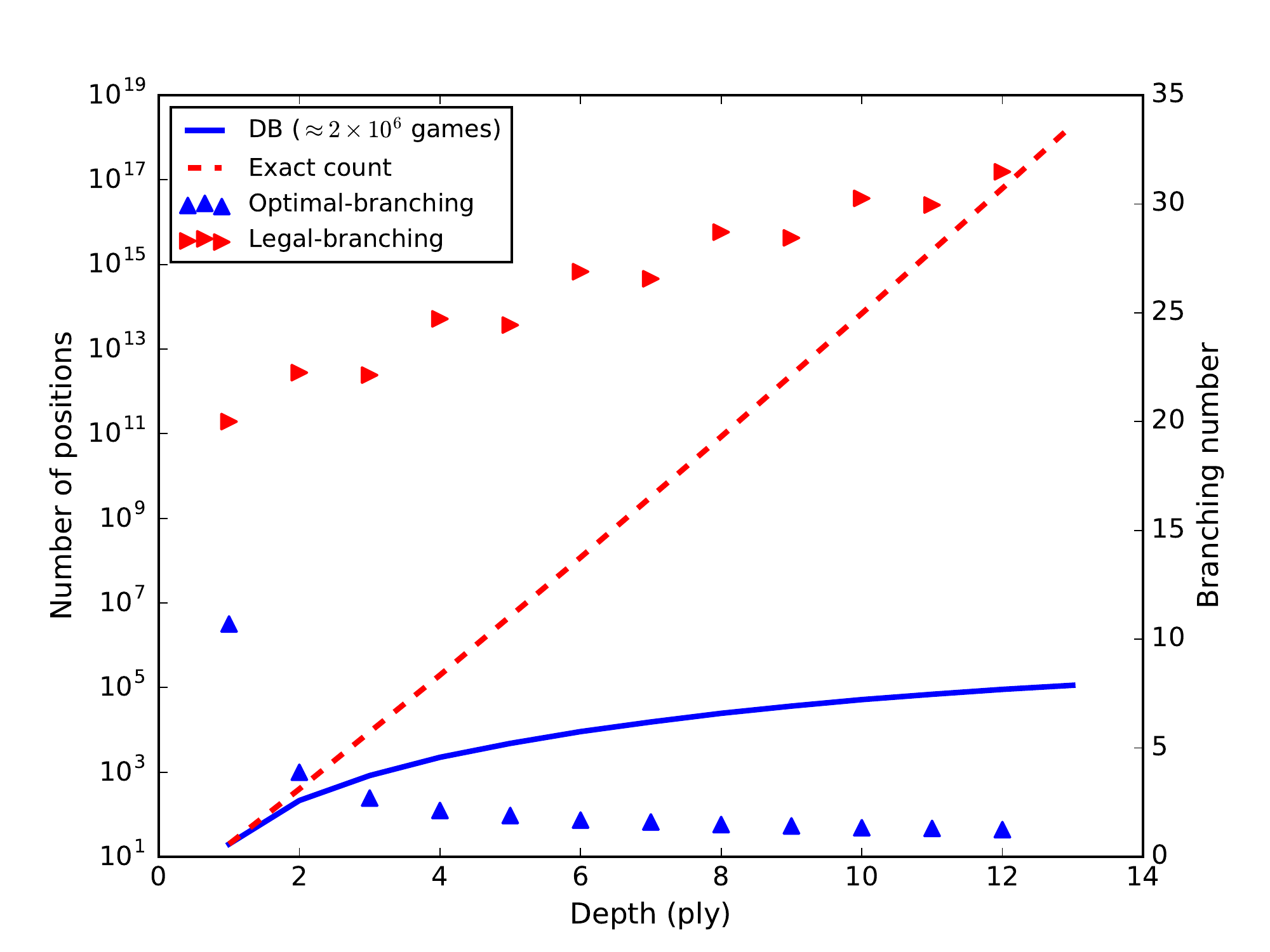}
\end{center}
\caption{\label{fig:poscount}
  Number of legal configurations accessible in chess at ply $t$ (dashed red line),
  and number of configurations found in a database (solid blue line) containing $2 \times 10^{6}$ games.
  Corresponding branching ratios are shown as symbols (right ordinate).
}
\end{figure}

The relative size of the pockets in state space can be estimated from
the path-length distribution by recurring to the theory of random graphs
\cite{Boccaletti2006,Newman2001,Newman2010}.
Let us view the game tree as an Erd\"os-Renyi random graph where essentially, any two nodes (states) are connected by one legal move with a
certain probability. Assuming that this probability is large enough so that
the connected component of the graph that we sample is strongly connected,
one expects that the average shortest path length between any two nodes in that
component scales as $\ell\sim\ln N/\ln z$, where $z$ is the
average branching rate. The size $N_2$ thus estimated from the large-distance
peak $\ell_2$ in the MC histogram can be viewed as an approximation
to $N_\text{accessible}$, the number of accessible chess states.

In the SPRES runs, the average branching rate is $z_1\approx23.8$
for pairs contributing to the peak at $\ell_1$, and $z_2\approx22.4$
for the larger-distance peak at $\ell_2$. SPRES also necessarily only
gives an upper bound for the shortest path length.
A chess-player's analysis  of some of the sampled paths (transposing between states by hand) indicates that this error is
about $10\%$, and up to about $20\%$ for the
pairs contributing to $\ell_2$.
Taking this into account, our estimate
is $N_\text{accessible}\approx\exp[35\ln22]\approx10^{47}$. The
pockets containing the actual games are estimated to have a relative
size $N_\text{blob}/N_\text{accessible}\approx\exp[(20\pm 2)\ln24]/\exp[35\ln22]
\approx10^{-20\pm 3}$.

Apart from endgame states with up to seven pieces, whose number is
known exactly (around $5\times10^{11}$ \cite{Lomonosov}),
only rough estimates exist regarding the size of the state space of chess,
and they all entail severe assumptions that do not even guarantee the
strict ordering $N_\text{possible}>N_\text{legal}>N_\text{accessible}$.
The most famous estimate is due to Shannon \cite{Shannon1950},
$N_\text{possible}\approx5\times10^{42}$ from a simple combinatorial
argument (uncorrected for captures and promotions).
The set of legal configurations is significantly smaller: by a factor of
about $10^{-7}$ under the approximations made by Shannon%
\footnote{Out of the $48!/(32!8!8!)$ possibilities to place eight white
and eight black pawns on the 48 squares between rank two and seven,
without captures only $15^8$ are accessible.}.
Including captures (but excluding promotions), an upper bound of
about $N_\text{legal}\approx2\times10^{40}$ has been shown recently
\cite{Steinerberger2015}, while an older calculation approximates
$N_\text{legal}\approx10^{50}$ including promotions \cite{upperbound}.
As of today, a reasonable estimate therefore continues to be
$N_\text{legal}=10^{42\pm7}$, with the ratio $N_\text{legal}/N_\text{possible}$
below a few percent.
In view of this, the value $N_\text{accessible}\approx10^{47}$
extracted from Fig.~\ref{fig:histogram} is entirely reasonable.
It is interesting to compare this state of uncertainty to the game of Go;
here, the number of legal states is known exactly
\cite{Tromp2007,A094777}, and the
question of reachability is presumably much less intricate,
although the numbers are much larger%
\footnote{In Go, $2.1\times10^{170}$ states are legal,
i.e., roughly $1\%$ out of all $3^{361}$ states.}.
This again points out that a major part of the complexity
of chess comes from the topology of its state space.

The branching numbers used in our estimate can be compared to
exact results known up to 13 plies \cite{Perft}.
Figure~\ref{fig:poscount} shows the number of configurations that can
be reached at a given ply as a function of plies (lines), and the
corresponding branching number (symbols). The latter approaches $z\approx35\pm5$ 
for the middle game, in agreement with the
estimate by Shannon \cite{Shannon1950}%
\footnote{Assuming a typical game length of 80 plies, this results in
the famous Shannon number for the game-tree complexity of chess,
$35^{80}\approx10^{123}$.}.
The values for $z$ found in SPRES are somewhat lower due to the fact
that we disqualify moves that obviously do not form part of connecting
pathways.

The number of optimal-play states is dramatically smaller than the number
of accessible states. It is also very difficult to estimate.
Taking the configurations encountered in our DB as a proxy (lower curve
in Fig.~\ref{fig:poscount}), this number seems to be about
$\mathcal O(10^6)$. (Note that the number of distinct configurations per
ply makes up less than $10\%$ of the size of the database, so that this number is probably not affected by the finite sample size.)
This and the fact that most DB configurations are effectively unconnected if one
restricts the connecting paths to near-optimal play,
suggest that the optimal-play nodes are well-separated sheets
that sparsely populate the pockets comprising the accessible state space.
Indeed, the branching number observed
in real games is only slightly above unity in the ply range used
to evaluate the histograms in Fig.~\ref{fig:histogram}: skilled human players do not
follow many branches in state space, but essentially make only one good move per
state.
The separation of the game into rather well separated branches occurs
during the opening moves.

%

\begin{figure}
\begin{center}
\includegraphics[width=1.0\linewidth]{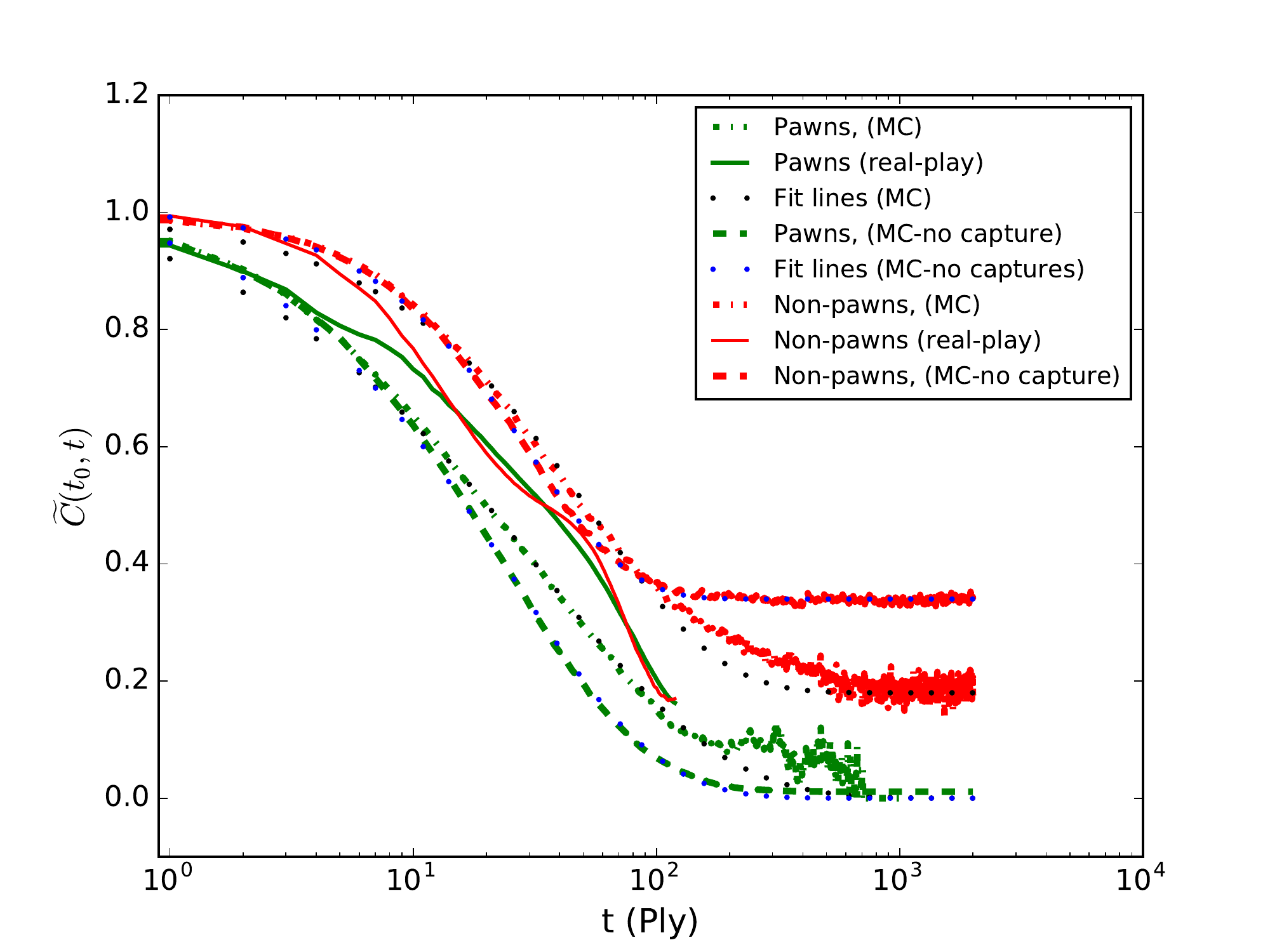}
\end{center}
\caption{\label{fig:corrnorm}
  Overlap correlation function, Eq.~\protect\eqref{eq:corr}, for pawns (green)
  and non-pawns (red) as a function of the number of plies starting from
  the initial chess configuration. Line styles distinguish different dynamics:
  real games from a database (solid), games of MC moves (dash-dotted),
  and games of MC moves without captures (dotted).
  Black dotted lines indicate stretched-exponential fits to the MC data.
}
\end{figure}

We now turn to the distinction of real-game play over MC dynamics
and explain the peculiar structure in state space discussed above.
One obvious difference is the inclusion of ``blunders'' in MC,
i.e., as we randomly pick allowed moves, we generate many very bad moves. These moves are typically not 
made by skilled players, and hence the real-game sheets 'disconnect'.

A more subtle difference leads to the splitting of state space into
several weakly connected pockets: the difference in pawn structure.
This can be seen in the overlap correlation function
\begin{equation}\label{eq:corr} C_{\alpha\beta}(t)=\frac1{\sqrt{N_\alpha(t)N_\beta(t)}}\left\langle
  \sum_{\substack{i=\text{a},\ldots\text{h}\\ j=1,\ldots 8}}n_\alpha(i,j,t)
  n_\beta(i,j,0)\right\rangle\,, \end{equation}
where $\alpha$ labels the type of piece,
$n_\alpha(i,j,t)$ is the occupation number of the labeled piece at
square $(i,j)$ and ply $t$, and $N_\alpha(t)$ the number of pieces
still on the board. For the following analysis, we group the pieces into 
pawns and non-pawns only.
Figure~\ref{fig:corrnorm} shows $C(t)$ extracted from 10000 DB games (using
the actual played trajectories),
2000 realizations of MC games and 2000 realizations of MC games without captures, with $t=0$ corresponding to the starting position of chess.

There is a striking difference between random play and real play in terms
of the correlation functions.
The real-game correlation functions display three regimes: (i) an initial decay
up to about ply~5, where both correlation functions follow those generated
by MC; (ii) a middle-game section between ply~5 and ply~50; and (iii)
a final decay after ply~50. These regimes match well with the distinction
between opening, middle game, and end game made in the heuristic
theory of chess developed by grandmasters \cite{Nimzowitsch}.
In the final regime, the real-game dynamics decorrelates much faster than
the MC dynamics, because human players eventually enforce the transition
to an end-game by a rapid exchange of pieces (``liquidation phase'').

In the middle-game, which is relevant for our discussion of the
state space, the most obvious difference
stems from the persistence of pawn--pawn correlations in real games.
In Fig.~\ref{fig:corrnorm}, the real-game correlation functions have a
bump between ply~5 and ply~20.
In contrast, the MC correlation functions are well described for all times by
stretched-exponential relaxation towards a long-time plateau, $C(t)\approx
f+(1-f)\exp[-(t/\tau)^\beta]$ with an exponent $\beta<1$. This reflects the
fact that the dynamics of chess pieces is highly collective, as typically,
the movement of any given piece is hindered by the others on the board
\cite{Klages2008}.  Under the MC dynamics, the pawns are more mobile than
the other pieces: the
fits yield characteristic decay times
$\tau_\text{pawn}\approx37.2\,\text{plies}$ and
$\tau_\text{non-pawn}\approx55.2\,\text{plies}$. The decay of the real-game dynamics
does not show this separation of time scales.

In real games, players tend to maintain a fixed pawn structure for much longer.
Keeping the pawn structure intact restricts moves to those
between configurations with larger overlap, and hence typically also shorter
path-length separations. The emphasis on pawn structure therefore prohibits
transpositions in real games.



\begin{figure}
\begin{center}
    \includegraphics[width=1.0\linewidth]{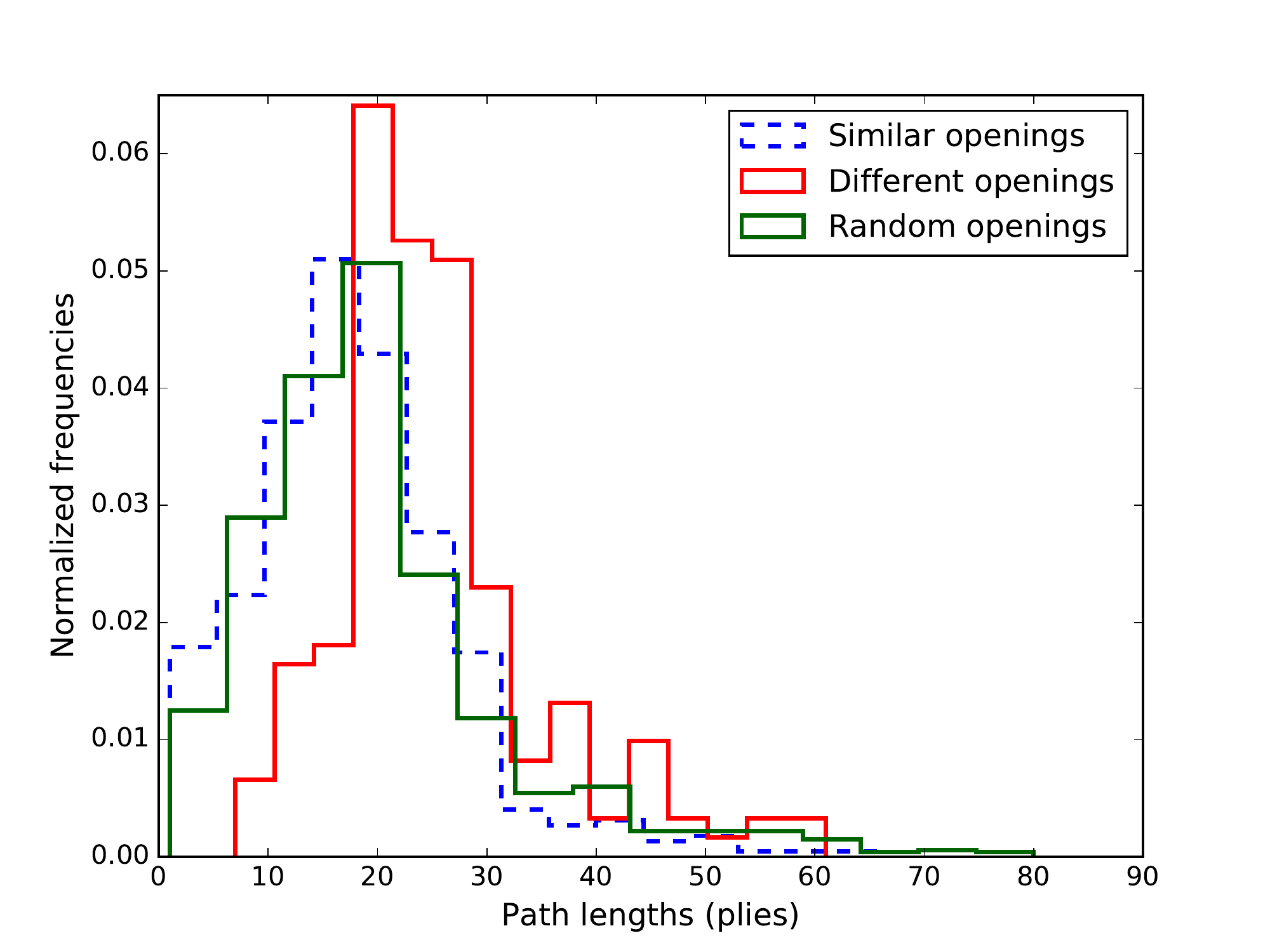}
\end{center}    
    \caption{\label{fig:opening_comparison} Distribution of path lengths between selected pairs of chess states sampled via SPRES, for database pairs that are classified according to the opening part of the game in which they arose. The red solid (blue dashed) histogram corresponds to pairs from different (similar) openings; see text for the classification. The green solid histogram repeats the DB results from Fig.~\protect\ref{fig:histogram} (no condition on the opening).
}
\end{figure}

To quantify the effect that the irreversible motion of the pawns has on the
blob structure of state space, we compare configurations drawn from
different openings.
Figure~\ref{fig:opening_comparison} shows the path length distributions of DB pairs
that were drawn from games with similar, different and random openings,
according to
their classification in the established chess-opening theory. The different
openings were selected to be sufficiently different regarding their initial
pawn moves. In particular, to obtain the similar-opening histogram,
we selected pairs where both configurations arose in so-called open games
(1.e4~e5), or both in closed games (1.d4~d5; 400 samples each). The
histogram for different openings was obtained by drawing one configuration
each from games with open and closed games (200 samples),
or one configuration each
from the Sicilian Defense (1.e4~c5) and the Indian Defence (1.d4~Nf6;
500 samples).

The path-length histogram for DB pairs from different openings displays
a noticable shift to larger distances: both the $\ell_1$ peak shifts to
larger values, and the tail around $\ell_2$ becomes more pronounced.
This supports our interpretation
that the pockets in state space correspond to different
openings with different pawn structures. As demonstrated by
Fig.~\ref{fig:opening_comparison}, two such pawn structures that divert
the game into farther separated sheets, are in particular those tied to
the ``open'' versus ``closed'' openings (1.e4~e5 versus 1.d4~d5).
Pairs from similar openings are, on the other hand,
easier to connect, because their pawn structure is compatible and allows
for transpositions.


In conclusion, we have applied SPRES sampling to
the problem of chess, using the resulting trajectories to infer the
topological structure of the game's state space.
Interestingly, SPRES even allows to make reasonable estimates regarding
the size of the state space, without referring to combinatorial
arguments.

Real games take place on well-separated ``thin sheets'' in this state space,
which are selected during the opening phase of the game, and dictated by the pawn structure.
Stretching the analogy to statistical physics, the real-game sheets in state
space are hypersurfaces of conserved pawn structure, and real games sample
the chess state space highly non-ergodically.
The branching number in these real-game sheets is only slightly above
unity, which loosely relates to a famous quote attributed to grandmaster
Emanuel Lasker, who stated that he thinks only one move ahead -- but a good
one.

It is quite remarkable that using SPRES, one can find ``reactive paths''
in the vast state space of chess. This should in principle offer a
dramatic improvement to computer chess programs, or various branches
of the mathematics of chess such as retrograde analysis \cite{Malikovic2014}
or the solution of chess puzzles \cite{Elkies2005}. Ordinary Monte Carlo
tree search (MCTS) is quite successful in Go
but performs poorly for chess, because
the state space of chess is highly fractured.
Employing methods from the computational physics of strongly
out-of-equilibrium systems offers an unexpected but promising take
on chess.

\begin{acknowledgments}
This project has been stimulated by COST Action MP1305 ``Flowing Matter'',
supported by COST (European Cooperation in Science and Technology).
We thank Mark Crowther for kindly providing us with the TWIC database of
chess games.
\end{acknowledgments}

\bibliographystyle{eplbib}
\bibliography{chess.bib}

\begin{thebibliography}{10}
\bibitem{config1}
{F}or the leftmost example, the shortest path found by SPRES was {1.Bg7 Bf1
  2.Nf6 d3 3.O-O d4 4.Bh6 Nc3 5.Bg7 g3}; this is nearly optimal.
\end{thebibliography}

\end{document}